\documentclass[preprint,12pt,authoryear]{elsarticle}

\usepackage{lineno,hyperref}
\modulolinenumbers[5]
\usepackage{algorithmic} 
\usepackage{algorithm}
\usepackage{graphicx}
\usepackage{multirow}    
\usepackage{hyperref}
\usepackage{geometry}
\usepackage{float}
\usepackage{subfloat} 

\usepackage{natbib}
\usepackage{amsmath}
\usepackage{url}
\usepackage{lineno,hyperref}
\modulolinenumbers[5]
\usepackage{color}
\usepackage{graphicx} 
\usepackage{subfigure} 

\date{}
\usepackage{amssymb}
\usepackage{natbib}
\setcitestyle{numbers,square}
\usepackage{threeparttable}

\usepackage{subfloat}
\usepackage{stfloats}

\usepackage{algorithmic}
\usepackage{algorithm}
\usepackage[algo2e,ruled,vlined]{algorithm2e}

\begin{document}

\begin{frontmatter}



\title{Object-Attentional Untargeted Adversarial Attack}


\author[1]{Chao Zhou}

\author[1]{Yuan-Gen Wang\corref{cor1}}
\cortext[cor1]{Corresponding author: }
\ead{wangyg@gzhu.edu.cn}

\author[2]{Guopu Zhu}

\address{\author{Chao Zhou}
	

}

\address[1]{School of Computer Science and Cyber Engineering, Guangzhou University, Guangzhou {\rm 510006}, China}
\address[2]{School of Computer Science and Technology, Harbin Institute of Technology, Harbin {\rm 150001}, China}

\begin{abstract}
Deep neural networks are facing severe threats from adversarial attacks. Most existing black-box attacks fool target model by generating either global perturbations or local patches. However, both global perturbations and local patches easily cause annoying visual artifacts in adversarial example. Compared with some smooth regions of an image, the object region generally has more edges and a more complex texture. Thus small perturbations on it will be more imperceptible. On the other hand, the object region is undoubtfully the decisive part of an image to classification tasks. Motivated by these two facts, we propose an object-attentional adversarial attack method for untargeted attack. Specifically, we first generate an object region by intersecting the object detection region from YOLOv4 with the salient object detection (SOD) region from HVPNet. Furthermore, we design an activation strategy to avoid the reaction caused by the incomplete SOD. Then, we perform an adversarial attack only on the detected object region by leveraging Simple Black-box Adversarial Attack (SimBA). To verify the proposed method, we create a unique dataset by extracting all the images containing the object defined by COCO from ImageNet-1K, named COCO-Reduced-ImageNet in this paper. Experimental results on ImageNet-1K and COCO-Reduced-ImageNet show that under various system settings, our method yields the adversarial example with better perceptual quality meanwhile saving the query budget up to 24.16\% compared to the state-of-the-art approaches including SimBA.

\end{abstract}




\end{frontmatter}


\section{Introduction}
Nowadays, deep convolutional neural networks (CNNs) have achieved great success in various computer vision tasks, such as image classification, object detection, and semantic segmentation. However, recent studies \cite{1}, \cite{2}, \cite{3} have shown that CNN-based classifiers could make wrong decisions about some images which contain imperceptible disturbances. These images crafted carefully by attackers are called adversarial examples \cite{2}. The existence of adversarial examples has threatened the applications of CNN models, especially for some security-sensitive fields. To overcome the security challenges from adversarial examples and improve the robustness of the model, many scholars have begun to pay attention to the research of adversarial attacks.\par

\begin{figure*}[tp]
	\centering
	\includegraphics[width=1\textwidth]{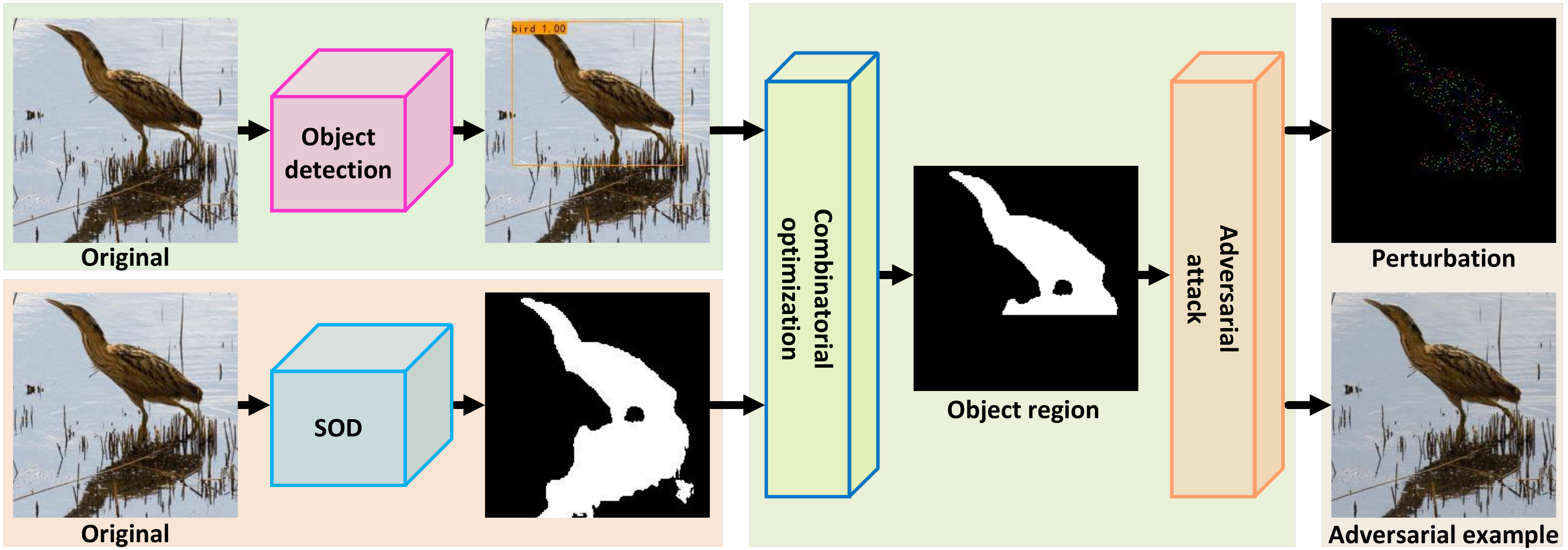}
	\vspace{-0.5cm}
	\caption{Overview of the proposed framework, which includes object detection, salient object detection, combinatorial optimization, and generation of adversarial example. }\label{fig1}
\end{figure*}

According to how much model information can be accessed, adversarial attacks are often divided into white-box attacks and black-box attacks. White-box attacks need to access the full parameters of the target model \cite{4}. On the contrary, black-box attacks can only get the prediction results of the model through the application interface. In practice, the parameters and intrinsic structure of the model are not public to users. Therefore, black-box attacks are more in line with the real-world application. In current black-box attacks, the importance of pixels is not distinguished. Thus, global pixel disturbance will be inevitable when generating adversarial examples. In fact, not of all pixels contribute equally to the CNNs. Studies have shown that the region corresponding to the ground truth has a major impact on the classifiers, while the background of an image cannot determine the classification result \cite{5}. For example, if an image is classified as ``cat'', the pixels in the object ``cat'' will play a key role. We call this set of pixels as ``object region". If we segment out the object region of an image and generate perturbation only on this region, the number of modified pixels can be significantly reduced and the attack success rate should be increased. Motivated by this, our work considers the importance of pixels in generating adversarial example. Next, we discuss the problem with the object segmentation.\par

The existing deep learning-based region detection methods mainly include object detection and salient object detection (SOD). As we know, YOLOv4 is a leading object detection method with high speed \cite{6}, and HVPNet is a lightweight SOD network which combines hierarchical visual perception (HVP) module with attention mechanism \cite{7}. However, YOLOv4 employs an anchor frame to detect an object. Hence, the detection result still contains a part of background except for the object region. On the other hand, the SOD region does not completely overlap with the object region. To sum up, neither of YOLOv4 and SOD is directly applicable to accurately segment an object. \par

\textbf{Why not use existing Semantic Segmentation algorithms?} As we know, semantic segmentation techniques can also segment the object region by classifying every pixel of an image. However, the classification of each pixel suffers from high computational cost and poor scalability \cite{5} \cite{17} \cite{18} \cite{19}. As for computational cost, training a semantic segmentation model is extremely hard due to pixel-level high-dimensional output and lack of high-quality dataset \cite{17} \cite{18} \cite{19}. On one hand, the semantic segmentation tasks require the pixel-level classification annotation, which however is much more labor-expensive and time-consuming. On the other hand, a new training task is necessary once any small changes are made in the convolutional layers. As for the scalability, it is extremely hard to extend the existing datasets to large-scale dataset with more categories since the dataset requires the pixel-level manual annotation \cite{17} \cite{18} \cite{19}. By contrast, the extension to object detection dataset is much easier since object annotation is much less labor. Besides, annotating a SOD dataset is relatively simple since the pixel-level classification is not needed. Therefore, it is not practical for our object segmentation task to use the existing semantic segmentation algorithms.

In this paper, we propose an object-attentional  adversarial example generation method for efficient untargeted attacks. To this end, we first present a new object segmentation algorithm  by integrating the advantages of YOLOv4 and HVPNet. Then we perform an adversarial attack only on the detected object region by leveraging Simple Black-box Adversarial Attack (SimBA). Extensive experimental results demonstrate the effectiveness of the proposed method. The major contributions of this paper can be summarized as follows:\par

\begin{itemize}
	\item We propose to perform adversarial attacks only on the object region of an image. Furthermore, an activation strategy is designed to avoid the reaction caused by the incomplete SOD, leading to an accurate object segmentation.
	
	\item We create a special dataset by extracting all the images containing the object defined by COCO from ImageNet-1K, named COCO-Reduced-ImageNet. Except for the ability to verify the proposed method, this dataset can be a supplement to the existing object detection datasets such as COCO.
	
	\item We test our method on four common models. The results show that under various system settings, our method yields the adversarial example with better perceptual quality meanwhile saving the query budget up to 24.16\% compared to the state-of-the-art approaches including SimBA.
	
\end{itemize}

\section{Related Works}
\subsection{Object Detection}

With the advances of CNNs, object detection algorithms have made a breakthrough. According to implementation process, the object detection can be mainly divided into two categories: two-stage methods and one-stage methods. The former is a type of the region-based CNN series algorithms, such as R-CNN \cite{8}, Fast R-CNN \cite{9}, Faster R-CNN \cite{10}, which are all implemented based on ``region proposal". In the two-stage methods, some typical algorithms such as selective search \cite{11} are first used to generate a candidate box at which an object may locate. Then the candidate box is classified and regressed. One-stage methods such as YOLOv4 \cite{6} and SSD \cite{12} can directly predict the categories and the locations of different objects by using an end-to-end network. Two-stage methods have higher accuracy but slower speed, while one-stage methods have lower accuracy but faster detection speed. In general, the black-box attacks require a large number of model queries for generating adversarial examples, which will consume high computational complexity. Based on this, our object segmentation algorithm selects YOLOv4. On the other hand, human visual system (HVS) has a strong ability of information processing, and can quickly capture more eye-catching area. The SOD technology aims to imitate the function of HVS. Thanks to the great progress of CNNs, the leading SOD methods have powerful capacity of feature representation \cite{13}, \cite{14}, \cite{15}, \cite{16}. However, most of these models call for a large amount of computational resources. For example, the recently proposed EGNet \cite{16} has 108M parameters. To improve the real-time performance of the model, a lightweight SOD network named HVPNet was proposed \cite{7}, which contains only 1.23M parameters. For saving computational resources, our object segmentation method chooses the lightweight HVPNet as the SOD network . \par

\subsection{Adversrial Attacks}
The concept of adversarial example was first proposed by Szegedy \textit{et al.}\cite{2} in 2013, which was usually applied in attacking image classification mdoels. The current research on adversarial attacks is mainly divided into two types: white-box attacks and black-box attacks. White-box attack is an attack implemented on the premise of fully knowing the parameters of target model. In \cite{20} Goodflow \textit{et al.} proposed a fast gradient sign method (FGSM). Carlini and Wagner \cite{21} for the first time dealed with adversarial attack problems from the perspective of optimization, and proposed an efficient white-box attack method which was named C\&W attack. The success rate of white-box attacks is high, but its practicability is not well because the internal parameters of the target model is often difficult to obtain in reality. Compared with white-box attacks, black-box attacks can perform attacks only by querying the prediction results of the target model, thereby it is more practical and more threatening to the target model.\par

The existing black-box attacks are mainly divided into gradient estimation methods and transferability attack methods. Gradient estimation methods estimate the gradient of an image in various ways, and then add adversarial perturbation to the direction of the estimated gradient. Chen \textit{et al.} \cite{22} proposed the Zeroth Order Optimization (ZOO) method to estimate the gradient of the target image. Cheng \textit{et al.} \cite{23} designed Opt-Attack method for hard label target networks. Tu \textit{et al.} \cite{24} proposed the framework AutoZOOM which uses an adaptive random gradient estimation strategy and dimension reduction technique to reduce attack queries. Ilyas \textit{et al.} \cite{25}  successfully exploited prior information about the gradient using bandit optimization. Transferability attack is based on the migratable property between different classification models. When the architecture and weight of the network are unknown, some attackers will choose to train another model from scratch, called the substitute model, so as to achieve their goal by performing white-box attack on the substitute model. Papernot \cite{26} was the first to propose to train a substitute model using the feedback data of query target model.\par

On the other hand, adversarial attacks to image classifiers can be divided into targeted attack and untargeted attack. Targeted attack requires the target model to classify the image into a specified error class, while untargeted attack only need to make the model classify the input image incorrectly. Finally, according to the different output forms, the target model can be divided into hard label model and soft label model. For the former, we can only get the information whether the input image is a certain category. For the latter, both the classification result and the probability values of each category can be obtained. In this paper, we focus on black-box untargeted attack for soft label scenario.\par

\section{Proposed Method}
Denote $x_0$ as the original image and input it to a pre-trained CNN classifier (target model). Then, we can obtain the output probability vector $p_n(x_0)$ where $n=1,...,N$ and $N$ denotes the number of categories of a classifier. Further, we can get the prediction category $\hat{c}(x_0)$ of the input image by
\begin{equation}
	\hat{c}(x_0) = \textrm{arg}\, \underset{n}{\textrm{max}}\, p_n(x_0).
\end{equation}
Denote $c(x_0)$ as the ground truth of the original image $x_0$. When $\hat{c}(x_0) = c(x_0)$, we say that the classifier realizes the correct classification of the image $x_0$. For untargeted attack setting, the goal of the adversarial attack is to find an adversarial example $x^*_a = x_0 + \delta $ subject to the following constraint
\begin{equation}
	x^*_a = \textrm{arg}\underset{x: \,\, \hat{c}(x_0)\neq c(x_0)}{\textrm{min}}\parallel x - x_0\parallel,
\end{equation}
where $\parallel\cdot\parallel$ denotes the distance metric. Usually, $L_2$-norm is used to measure the distortion of the adversarial example. The smaller the distance, the smaller the distortion.

This paper considers the adversarial attack only on the object region, which can reduce the number of attacked pixels, thereby decrease the distortion, and in turn increase the concealment of the perturbation. Fig 1 shows the overall framework of our attack method. It includes object detection, salient object detection, combinatorial optimization, and generation of adversarial example. In what follows, we will describe each of them in detail.

\begin{figure*}[tp]
	\centering	\includegraphics[width=1\textwidth]{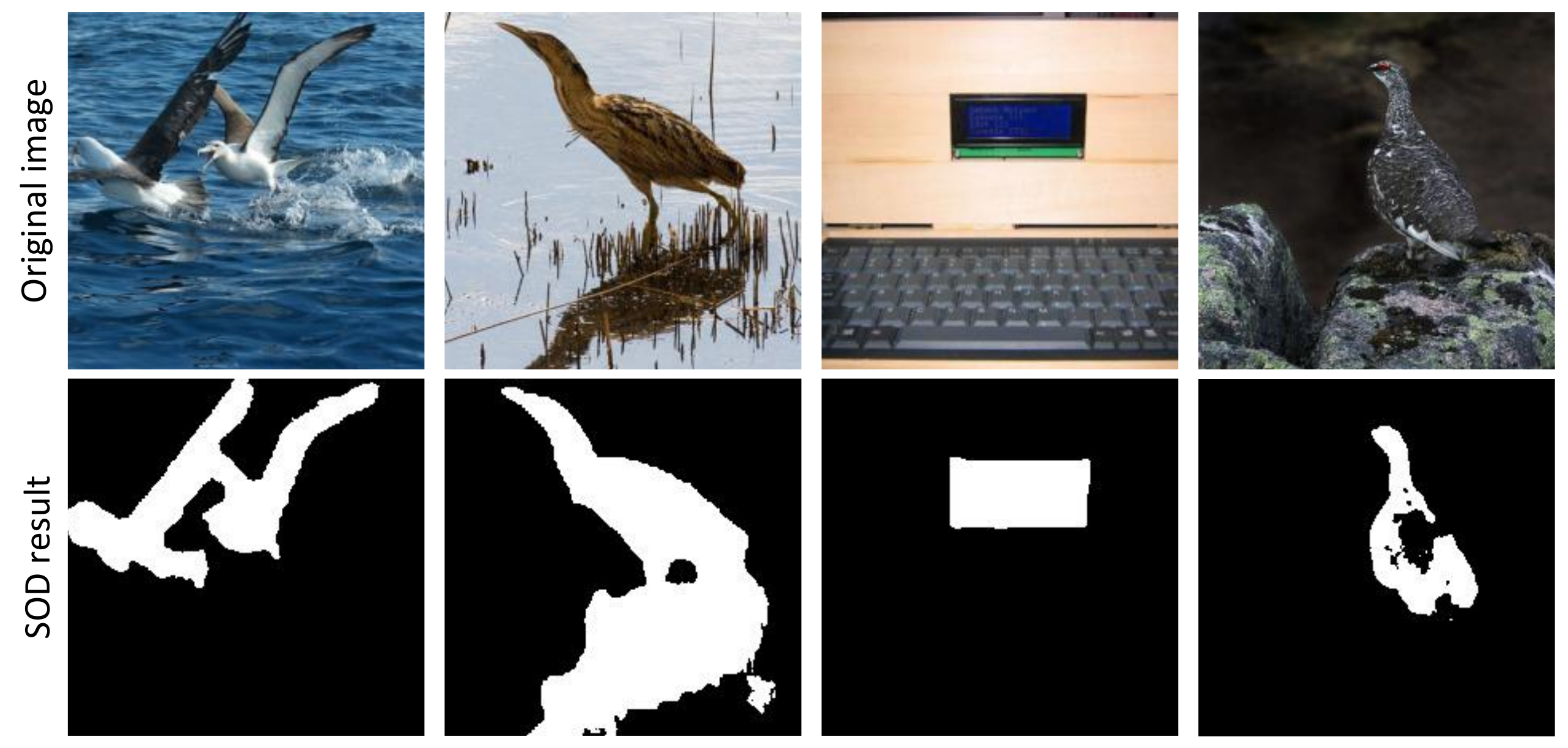}
	\caption{Illustration of SOD result. Here, white parts in the second row represent salient region.}
\end{figure*}

\subsection{Object Region Detection}
Different from image classification task, object detection not only recognizes what the object is, but also needs to locate where the object is. When an object is detected in the image, we need to circle the object with a frame and annotate the corresponding confidence value. Since current object detection models can quickly locate the object in an image, we utilize existing models to estimate the object region. YOLOv4 is a one-stage object detection method and has a very fast detection speed. For example, for an input image with $416\times 416\times3$ size, the processing speed under Tesla V100 can reach 65 fps. The test result on the COCO dataset \cite{27} reaches 43.5$\%$ average precision (AP) (65.7$\%$ AP with 0.5 threshold). 

Based on the advantages of YOLOv4, our method employs the pre-trained YOLOv4 model on the COCO dataset for the object detection (short for YOLOv4 thereinafter for simplicity). Denote $f_1$ as the output function of YOLOv4, $S_1$ as its regional detection result, and $P$ as the corresponding confidence of the detected object. Then we have
\begin{equation}
	[P,\, S_1] = f_1(x_0) \,\,\,\, \textrm{s.t.}\, P > P_t,
\end{equation}
where $P_t$ is the object output threshold which is set empirically. Only when the confidence of an object is greater than $P_t$,  this object will be output by the detector. Obviously, different values of $P_t$ will affect the AP value of object detection result.

\subsection{Salient Object Detection}
HVS can quickly locate region of interest (ROI), then only processes the ROI and tends to ignore the other area. This visual information processing is called ``visual attention mechanism''. Because the attention mechanism can quickly lock the ROI in the visual scene, it can greatly reduce the amount of data computation and accelerate the speed of information processing. This is very attractive for the machine vision applications with limited computing resources and high real-time requirements. 

Built upon the above characteristics, salient object detection (SOD) has developed to model HVS. The processing pipeline of HVS is in a hierarchical structure. Multiscale visual signals are hierarchically processed in different cortex areas that have different population receptive fields (PRF) \cite{28}. Inspired by this, Liu \textit{et al.} proposed an HVP model to simulate the structure of primate visual cortex, which can be expressed by
\begin{equation}
	R_r(x_0)=\left\{
	\begin{array}{cl}
		F^{1\times1}(x_0), &  \textrm{if} \,\, r = 1 \\
		F^{1\times1}(\hat{F}^{3\times3}_r(F^{1\times1}(\hat{F}_{1}^{r\times r}(x_0))), &  \textrm{if} \,\, r > 1, \\
	\end{array} \right.
\end{equation}
where $F^{1\times 1}$, $\hat{F}_{1}^{r\times r}$, and $\hat{F}^{3\times3}_r$ are the vanilla convolution with the kernel size of $1\times 1$, DSConv \cite{29} with the kernel size of $r\times r$, and DSConv with the kernel size of $3\times 3$ and the dilation rate of $r$.
Furthermore, a lightweight SOD network termed HVPNet was designed by combining HVP module and attention mechanism. HVPNet has only 1.23M parameters and reaches the running speed of 333.2 fps (336$\times$336$\times$3 frame size). Thanks to these advantages of HVPNet, we select it to detect the SOD of an image. Denote $f_2$ and  $S_2$ as the function and output of HVPNet, respectively. Thus, we have 
\begin{equation}
	S_2 = f_2(x_0).
\end{equation}
The SOD output of HVPNet is in the form of binary images. An example is shown in Fig 2, where the white part represents the salient object and the black part represents the background. We can see from Fig 2 that the salient object region  can be well detected by HVPNet.\par

\begin{figure*}[tp]
	\centering	\includegraphics[width=1\textwidth]{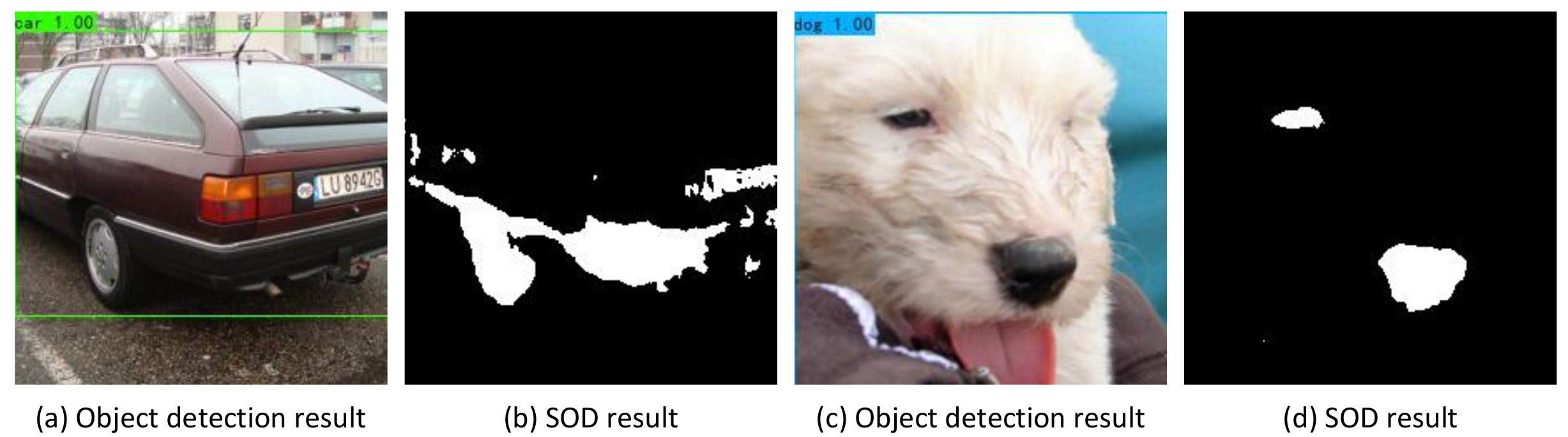}
	\caption{Incompleteness phenomena appears in the SOD result: the intersection of object detection result and SOD result is much smaller than the object region.}
\end{figure*}

\subsection{Combinatorial Optimization}
It is known that YOLOv4 has the advantage to quickly locate the object in an image and HVPNet can accurately detect the salient region of an image. But they have their respective limitations when used to estimate the object region separately. The anchor-box of YOLOv4 usually contains some background region, which is not important in classification tasks. That means YOLOv4 cannot well segment out a real object. The salient region from HVPNet is not always overlapped with the object region. For example, as shown in the second column of Fig 2, the reflection of bird is detected as a salient region, but not as a real object. Besides, as shown in the third column of Fig 2, the classification ground truth of the original image is keyboard, but the salient region detected by HVPNet is the LCD screen. In this case, if the SOD is uniquely used to determine the object region, we cannot obtain the real object. This will negatively impact the performance of the object segmentation. \par

Taking the above two issues into consideration, we propose a combinatorial optimization strategy to calculate out the real object region. In most case, the salient region of an image overlaps with its object region. Therefore, we propose to compute the intersection between the detection results of YOLOv4 and HVPNet as the object region. On the other hand, when the SOD result is not contained in YOLOv4 detection region, we only consider the YOLOv4 detection result as object region. This is because YOLOv4 has the strong ability of detecting the objects. By this, we can successfully solve the above two limitations and leverage the advantages of both detectors.

By our experiments, we find that a detection incompleteness phenomena appears in the SOD result. That is the SOD region does not contain the whole object. As shown in Fig 3, the intersection is much smaller than the real object region (car and dog). To overcome this problem, we design a salient region activation factor $k$ to further improve the combinatorial optimization strategy. Denote $S_1$ as the object detection result from YOLOv4, for example, the region in the box of Fig 3(a) and (c). Similarly, denote $S_2$ as the SOD result, such as the white region of Fig 3(b) and (d). Thus we can obtain the intersection $S=S_1\cap S_2$. Obviously, the intersection $S$ is only a small part of the whole object (car or dog). In this case, if $S$ is taken as the object region to be attacked, the number of queries will increase significantly and the attacked region will have a very low perceptual quality. Although YOLOv4 cannot accurately detect the object's contour, it can contain the whole object with smallest area. Based on this observation, we propose a salient detection activation strategy to solve the incomplete detection problem. The activation strategy is controlled by an activation factor $k$, which is calculated by 
\begin{equation}
	k = \frac{S_1}{S_1\cap S_2}.
\end{equation} 
When $k>\epsilon$ ($\epsilon$ is a super parameter and will be set empirically), the detection incompleteness occurs. In this case, only the YOLOv4 detection result is used as the estimation of the object region while the SOD result is discarded. According to our experiments,  a good value of $\epsilon$ will significantly improve the attack efficiency. \par

\subsection{Generate Adversarial Example}
We adopt the SimBA algorithm to generate adversarial example. SimBA randomly search each pixel of the whole image space to perform an attack in a simple and brute manner. Our method improves SimBA by restricting the search to the detected object region. Denote $I$ with $M\times M$ dimensions as an identity matrix where $M$ is the total number of pixels of an input image. Our method first constructs an orthogonal set $Q$ by picking all object row vectors from $I$, where the object row vector refers to as these vectors whose nonzero element positions correspond to the detected object region (Note that the input image is also flattened to a vector). The construction of $Q$ can be defined by 
\begin{equation}
	Q = I\odot V,
\end{equation} 
where $\odot$ denotes element-wise dot product and $V$ is a matrix obtained by copying the flattened image vector (in which the detected object region takes 1 and the other region takes zero) $m$ times.

\begin{algorithm2e}[htb]
	\caption{Object-Attentional Untargeted Adversarial Attack Algorithm}
	\label{alg3}
	\begin{algorithmic}[1]
		\REQUIRE image $x_0$ and the ground truth $c(x_0)$, object detection model $f_1$, SOD model $f_2$, maximum query number $N$, salient region activation threshold $\epsilon$, super parameter $P_t$.
		\ENSURE  adversarial example   $x_a$
		\STATE {$i = 0 $, $x_a=x_0$ }	 		
		\STATE $p = \textrm{max} (p_n(x_0)) $ 
		\STATE {Construct an initial $Q$:  $Q=\textrm{construct}(x_0,f_1,f_2,\epsilon,P_t)$}  
		
		\WHILE{ $\hat{c}(x_a) = c(x_0) $ and $i < N$}
		\STATE {Randomly select an orthogonal vector $q$ from $Q$} 	
		\FOR{$\alpha \in \{\mu, -\mu\} $ } 
		\STATE $p' = \textrm{max}(p_n(x_a + \alpha q))$ 
		\IF{$p' < p$}
		\STATE $x_a = x_a + \alpha q$
		\STATE $p' = p $				
		\STATE break
		\ENDIF
		\ENDFOR 
		\STATE $Q=Q\backslash q$, ``$\backslash$'' denotes the deletion operation	
		\STATE $i++$	
		\ENDWHILE
		\STATE \Return adversarial example $x_a$		
	\end{algorithmic}
\end{algorithm2e}

Then, our search attack can focus on the detected object region. In each iteration, we randomly select a vector $q$ from $Q$ ($q\in Q$). Our attack takes $x_a\,=\,x_0\,+\,\mu q$, where $\mu$ is the disturbance step size. If the prediction probability of the correctly-classified image $p(y|x_a)$ is reduced, we add the disturbance to the target image in this step. Otherwise, let $x_a=x_0-\mu q$. The above process is repeated by re-picking a direction $q$ from $Q$ on the basis of the previous attack result, and the cycle continues until the attack successes or the given maximum number of queries is reached. Algorithm 1 gives the complete attack process of our method. Since our attack is implemented on the object region of the image, the search number of $q$ will be greatly reduced, thereby significantly improving the attack efficiency. The detailed process of constructing $Q$ and selecting  $q$ is summarized in Algorithm 2.\par

\begin{algorithm2e}[ht]
	\caption{Construct the orthogonal set Q Algorithm}
	\label{alg3}
	\begin{algorithmic}[1]
		\REQUIRE image $x_0$, $M$ denotes the number of pixels of $x_0$,  object detection model $f_1$, SOD model $f_2$, salient region activation threshold $\epsilon$, super parameter $P_t$, image size [$w$, $h$]
		\ENSURE  orthogonal set $Q$
		
		\STATE $I = \textrm{IdentityMatrix}(M)$, $S=\textrm{ones}(w,h,3)$, $S_1=S_2= \textrm{zeros}(w,h)$   
		\STATE Get object detection result $(l,r,t,b) \gets f_1(x_0,P_t) $ 
		\STATE $S_1(t:b,l:r)= 1$      
		\STATE Get SOD result $S_2 \gets f_2(x_0) $  
		\STATE$\hat{S} = S_1\odot S_2 $  \% $\odot$ denotes element-wise dot product
		
		\IF{$\frac{\textrm{sum}(S_1==1)}{\textrm{sum}(\hat{S}==1)} > \epsilon $}
		\STATE $\hat{S}=S_1$
		\ENDIF 
		
		\STATE $S=S \odot \hat{S}$
		\STATE $v = \textrm{Reshape}(S) $ \% Reshape $S$ to a column vector
		\STATE $V=\textrm{Repmat}(v, M)$ \% Repeat the column vector $v$ in $M$ times
		\STATE $Q = I\odot V$
		\STATE $Q = \textrm{DeleteZeroRow}(Q)$ \% Delete all-zero rows from $Q$
		\STATE \Return $Q$  
	\end{algorithmic}
\end{algorithm2e}

\section{Experimental Results}
\subsection{Datasets and Target Models}
Two datasets are used for the evaluation. One is the validation set of ImageNet-1K. This dataset covers most of the object categories that could be seen in daily life, and each image has been manually labeled with a category. ImageNet-1K has 1,000 categories, and each category in the validation set contains 50 images. We know that COCO contains 80 object categories most of which are included in the categories of ImageNet-1K. This dataset is the most widely used in the field of object detection so far. In order to validate the effectiveness of the proposed object-attentional mechanism, we construct a special dataset, called COCO-Reduced-ImageNet-1K, which is obtained by eliminating all the images of ImageNet-1K validate set whose labels do not belong to the category of COCO dataset. Finally, COCO-Reduced-ImageNet-1K contains 298 categories, each of which has the same 50 images as that of the ImageNet-1K validation set.

In the experiments, YOLOv4 model pre-trained on COCO dataset is used for the object detection, and HVPNet model pre-trained on DUTS dataset \cite{30} is used for the SOD. Four benchmark target models are selected for adversarial attack, which are ResNet50, ResNet18, VGG16, and Inception V3. 

\subsection{Metrics and Parameters Selection}
Five metrics are adopted to evaluate the black-box adversarial attacks: 1) Success rate (the possibility that the attacker can successfully find the adversarial example); 2) Average queries (how many queries were required on average); 3) Median queries (how many queries are most common); 4) Average $L_2$-norm (how much distortion causes on average ); 5) Median $L_2$-norm (how much distortion is common). Clearly, the fewer the average queries (or median queries), the better performance. Similarly, a smaller average $L_2$ distortion (or median $L_2$ distortion) indicates that the perturbation produced by the adversarial attack method is more invisible, and a higher attack success rate shows that the method has stronger attack ability.\par

\begin{figure*}[tp]
	\centering	\includegraphics[width=1\textwidth]{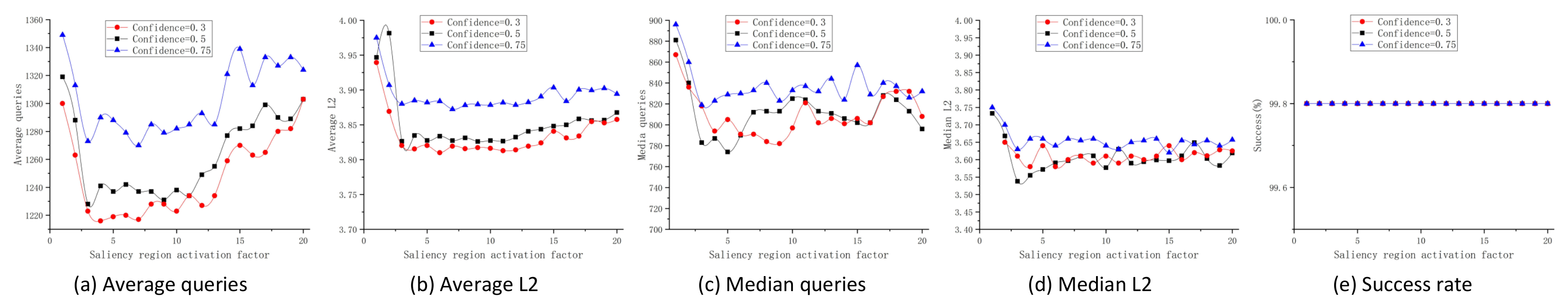}
	\caption{Attack statistic curves using different super parameters $P_T$ and $\epsilon$. Three values for $P_T$ are tested on COCO-Reduced-ImageNet dataset and ResNet50.}
\end{figure*}

According to the experiments, we find that the super parameters $P_t$ and $\epsilon$ can directly affect the estimation accuracy of object region, and then decide the attack efficiency. We test three values for $P_t$, which are 0.3, 0.5 and 0.75, respectively. For each $P_t$ value, we take 20 values of $\epsilon$ within $[1, 20]$ with a unit interval, and then compute the attack performance. The results of all the combinations are shown in Fig 4. There are three curves corresponding to three different values of $P_t$. It can be seen from Fig 4 that when the value of $P_t$ is small, the average queries, average $L_2$-norm, and median queries of the model are relatively low, while the difference of the median $L_2$-norm is not obvious, and the attack success rate has no change (see red curve). Based on this experiment, we take a low value of $P_t$. This is because appropriately reducing the confidence threshold of the object output enables more potential object region to be detected. By observation, we find that it is the most appropriate to take the value of $\epsilon$ in $[3, 10]$. If the value of $\epsilon$ is too large or too small, the attack effect will become worse. In the experiment, the optimal attack result on COCO-Reduced-ImageNet is obtained when $P_t =0.3$ and $\epsilon =3$, where the average queries is 1,216, the Median queries is 794, the average $L_2$-norm is 3.82, and the average $L_2$-norm is 3.58. Therefore, our method takes $P_t =0.3$ and $\epsilon =3$ in all the experiments. We extract 1,000 images from each dataset as the target images, which are all required to be correctly classified by the target model. In Algorithm 1, the maximum number of queries allowed is set to 20,000 ($N=20,000$).

We compare our method against three leading untargeted black-box attack algorithms, which are NES \cite{31}, Bandits-TD \cite{32}, and SimBA \cite{33}. In addition, we perform an ablation study to verify the contribution of object detection used by YOLOv4 only (SLY), salient object detection used by HVPNet only (SLH), and object segmentation used by combining YOLOv4 with HVPNet (OA). Note that the normalized disturbance step $\mu$ of SimBA, SLY, SLH and OA are set to a fixed value of 0.2. In the following, we present each experimental result and their analysis in detail.

\begin{table*}
	\caption{Statistical data of attacks for ImageNet-1K. The best competitor is highlighted in bold.}
	\label{tab:freq}
	\resizebox{\textwidth}{!}
	{
		\setlength\tabcolsep{2pt}
		\begin{threeparttable}
			\begin{tabular}{|c|c|c|c|c|c|c|c|c|c|c|c|c|c|c|c|c|c|c|c|c|}
				\hline
				\multirow{2}{*}{Attack} & 
				\multicolumn{4}{c|}{Average queries} & \multicolumn{4}{c}{Median queries} &\multicolumn{4}{|c|}{Average $L_2$-norm}  & \multicolumn{4}{c|}{Median $L_2$-norm } &  \multicolumn{4}{c|}{Success rate ($\%$)} \\
				\cline{2-21}
				
				& N50& N18 &G16 &V3 & N50& N18 &G16 &V3& N50& N18 &G16 &V3& N50& N18 &G16 &V3& N50& N18 &G16 &V3\\
				\hline		
				NES \cite{31} & 3615&2801 &2234 &7714&2150 & 1720 &1280 &3830&4.12 &3.86 & \textbf{3.17}& \textbf{4.20}&4.30    & 3.86&     \textbf{3.29}& 5.00&95.54 & 97.35&96.26 &77.26  \\
				\hline
				TD \cite{32}& 2104 &1445&3308 &7101 &\textbf{882} &\textbf{538}&896 &2349 &4.89&4.92 &4.74 &4.91&5.00 & 5.00&5.00&\textbf{5.00}&97.00 &98.18 &91.78&73.90  \\
				\hline
				BA \cite{33}& 1538 &1194 &1706&3785 &1018&873  &852&1988&4.21&3.95 &3.82&6.52 &3.95&3.81 & 3.65&5.80 &99.70 &99.70 &97.39&\textbf{94.90}  \\
				\hline
				SLY (our) & 1491 &1124 &1603&4079 &966 &801  &768&2198&4.08&3.85 &3.69&6.45 &3.84&3.75 & 3.46&5.92 &99.70 &\textbf{99.80} &\textbf{97.59}&93.40  \\
				\hline
				SLH (our)& 1675 &1197 &1627&3687 &950 &777  &739&1741&4.20&3.86 &3.65&6.11 &3.92&3.64 & 3.36&5.49 &99.20 &99.70 &97.00&94.80  \\
				\hline
				OA (our) &\textbf{1465}&\textbf{1118} &\textbf{1559}&\textbf{3573} &916&759 &\textbf{702} &\textbf{1790}  &\textbf{4.04}&\textbf{3.80} &3.63 &6.39 &\textbf{3.81}&\textbf{3.64} &3.36 &5.59 &\textbf{99.70}& 99.70 &97.50 &94.70  \\       
				\hline
			\end{tabular}
			\begin{tablenotes}
				\footnotesize
				\item Abbreviated by TD: Bandits-TD, BA: SimBA, N50: ResNet50, N18: ResNet18, G16: VGG16, and V3: Inception V3.
			\end{tablenotes}
		\end{threeparttable}
		
	}
\end{table*}

\subsection{Experimental Results on ImageNet-1K}
Table 1 shows the experimental results on the ImageNet-1K dataset. From Table 1, we can see that compared with  the four common deep learning networks, our method uses smaller average queries and median queries. Especially for the target network VGG16, our method reduces $8.62\%$ and $17.61\%$ of SimBA in terms of the average and median queries. And for $L_2$-norm, our method generally has a great improvement in both average and median $L_2$-norm (namely $4.97\%$ and $7.95\%$ respectively). In terms of attack success rate, our method remains the same level as the SimBA method. This is due to the fact that our object segmentation is accurate and adversarial attack on the object region is effective. We can also observe from the last three rows of Table 1 that both SLY and SLH can improve the baseline method (SimBA) in terms of query number and distortion, but perform worse than the OA method. This ablation study further validates the effectiveness of the proposed method. That is, the object detection by YOLOv4 is only a rectangular region which is not accurate enough for a real object. Although the SOD method has advantage in salient region detection, there exist many detection errors and detection incompleteness. Our method takes full advantage of these two detection methods and designs an activation strategy, thereby boosting the attack performance significantly.\par

\subsection{Experimental Results on COCO-Reduced-ImageNet}
Table 2 shows the experimental results on the COCO-Reduced-ImageNet dataset with the same parameter setting. From Table 2, we can see that our method still achieves smaller average queries and median queries, which is better than the performance on the ImageNet-1K dataset. For VGG16, our method reduces $18.97\%$ and $24.16\%$ compared with the baseline SimBA method. In terms of average $L_2$-norm and median $L_2$-norm, our method reduces $12.37\%$ and $14.77\%$ of SimBA, which is much better than the ImageNet-1K dataset. In the other hand, for Inception V3, our method even has a higher attack success rate. In addition, the results in Tables 1 and 2 indicate that Inception V3 has a higher defense capability against our attack method, followed by VGG16. ResNet18 is the weakest against our method.\par

\begin{table*}
		\caption{Statistical data of attacks for COCO-Reduced-ImageNet. The best competitor is highlighted in bold.}
		\label{analysis}
			\resizebox{\textwidth}{!}
			{
				\setlength\tabcolsep{2pt}
				\begin{threeparttable}		
					\begin{tabular}{|c|c|c|c|c|c|c|c|c|c|c|c|c|c|c|c|c|c|c|c|c|}
						\hline
						\multirow{2}{*}{Attack} & 
						\multicolumn{4}{c|}{Average queries} & \multicolumn{4}{c|}{Median queries} &\multicolumn{4}{|c|}{Average $L_2$-norm}  & \multicolumn{4}{c|}{Median $L_2$-norm } &  \multicolumn{4}{c|}{Success rate ($\%$)} \\
						\cline{2-21}
						
						& N50 &N18 &G16 &V3 & N50&N18 &G16 &V3& N50&N18 &G16 &V3& N50&N18 &G16 &V3& N50&N18 &G16 &V3\\
						\hline		
						NES\cite{31} &3080 &2627 &1788 &6960 &1920 &1660 &1150 &3230 &3.82 &3.71 &\textbf{3.10} &4.20 &4.12 &3.84 &\textbf{3.12}&5.00 &97.36 &97.54 &97.83 &79.96  \\
						\hline	
						TD \cite{32}&1621&2309&2750 &5746 &\textbf{732}&\textbf{494} &812 &\textbf{1568}&4.93 &4.63&4.78 &\textbf{4.98} &5.00 &5.00 &5.00 &\textbf{5.00} &98.47 &92.13 &93.34 &81.50 \\
						\hline	
						BA \cite{33}&1462&1161 &1502 &3896 &1031&849 &923 &1921 &4.17&3.96 &3.96 &6.60 &4.00&3.84 &3.86 &5.76 &99.80&99.90 &98.59 &94.40  \\
						\hline	
						SLY (our)	&1319&1066 &1345 &3985 &881 &774 &764 &2029 &3.94&3.80 &3.69 &6.42 &3.73&3.64 &3.51 &5.78 &99.80&99.90 &98.59 &93.70 \\
						\hline	
						SLH (our)&1354&1172 &1312 &3521 &868	&701 &707 &1691 &3.92&3.82 &3.56 &6.00 &3.68&3.57 &3.36 &5.36 &99.80&99.90 &98.59 &95.10\\	
						\hline	
						OA (our) &\textbf{1216}&\textbf{1012} &\textbf{1217} &\textbf{3243} &794&702 &\textbf{700} &1625&\textbf{3.82}&\textbf{3.69} & 3.47&6.18 &\textbf{3.58}&\textbf{3.51} &3.29 &5.43 &\textbf{99.80}&\textbf{99.90} &\textbf{98.59} &\textbf{96.20}  \\		  				
						\hline
					\end{tabular}
					\begin{tablenotes}
						\footnotesize
						\item Abbreviated by TD: Bandits-TD, BA: SimBA, N50: ResNet50, N18: ResNet18, G16: VGG16, and V3: Inception V3.
					\end{tablenotes}
				\end{threeparttable}					
			}
	\end{table*}

Fig 5 shows the distribution histogram of the number of queries required for successful attack over 1,000 random images of the COCO-Reduced-ImageNet dataset. We take the increment 200 as an interval and count the number of images whose queries are within this interval. Images with more than 5,000 queries and failed attack examples are counted as an interval since most images can be attacked successfully within 5,000 queries. It is obvious that the query distribution of the four models is generally biased to the left. In the interval of low queries, the number of images of method SYL, SLH and OA is higher than that of SimBA method, while the opposite is true for the right. For example, for ResNet18, the numbers of the images that can be successfully attacked within 200-400 queries are as: SimBA is 125, SLY is 136, SLH is 153, and OA is 155. While the statistical data of 2600-2800 queries are: SimBA is 20, SLY is 15, SLH is 16, and OA is 14. This query distribution clearly demonstrates that the majority of images can be successfully attacked with less queries by our method. \par

\begin{figure*}[tp]
	\centering	\includegraphics[width=1\textwidth]{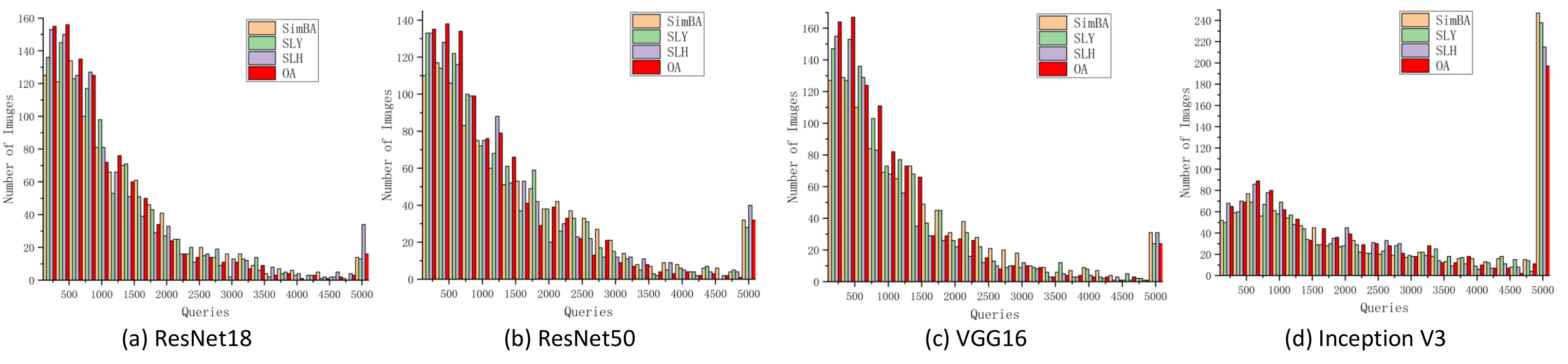}
	\caption{Distribution of the number of successfully attacked images with the model queries.}
\end{figure*}

\subsection{Visual results}
Our method performs attacks only on the pixels of the object region where the gradient change is usually larger than that of the smooth background. Thus, it will less likely to be detected by the human eyes when small disturbance is added to the areas with sharp gradient change. Two metrics are adopted to assessment the image quality: 1) PSNR (Peak Signal-to-Noise Ratio); 2) SSIM (Structural Similarity). Both PSNR and SSIM are used to calculate the difference of two images. The higher the SSIM (or PSNR), the more similar the adversarial example and the original image are. We average the PSNR and SSIM on 1000 samples. Table 3 shows the experimental results on the ImageNet-1K and COCO-Reduced-ImageNet datasets. Obviously, compared with SimBA, our method obtains better performance in terms of both PSNR and SSIM.
Especially, for Inception V3 on the COCO-Reduced-ImageNet dataset, our method improves the PSNR by 1.2329 and the SSIM by 0.0096 respectively, compared with the baseline (SimBA). \par

Fig 6 shows the visual effect of four attack methods of SLY, SLH, OA and baseline SimBA on ResNet50 and COCO-Reduced-ImageNet. The columns 1 and 4 show the original image, the second and fifth columns show the visual restoration of adversarial perturbation, and the third and last column show the adversarial examples generated by the corresponding methods. It can be seen from the second and fifth columns that our attack method successfully reduces the range of adversarial perturbation and focuses on the object region. Specifically speaking, the perturbation produced by SLY is limited in the rectangular region. This is because YOLOv4 only segments out a rectangular object. Although the perturbation in subfigure (row 3, column 5) is not limited to rectangular, a part of perturbation is added to the region of ``Bird'' reflection due to the detection error of SOD. When carefully observing the third and last columns, we can find that our attack method (OA) has better visual effect because the smooth area of the image is well preserved. Therefore, the perturbations generated by our method are more imperceptible.

\begin{table*}
	\caption{PSNR and SSIM results. The best competitor is highlighted in bold.}
	\label{analysis}
		\resizebox{\textwidth}{!}
		{
			\setlength\tabcolsep{2pt}
			\begin{threeparttable}		
				\begin{tabular}{|c|c|c|c|c|c|c|c|c|c|}
					\hline
					\multirow{2}{*}{dataset} &
					\multirow{2}{*}{Attack} & 
					\multicolumn{4}{c|}{PSNR} & \multicolumn{4}{c|}{SSIM} \\
					\cline{3-10}
					
					& &ResNet50 &ResNet18 &VGG16 &Inception V3 & ResNet50&ResNet18 &VGG16 &Inception V3\\
					\hline		
					\multirow{2}{*}{ImageNet-1k} &SimBA &37.3212 &37.5309 &38.1342 &37.3963 &0.9772 &0.9785 &0.9792 &0.9698  \\
					\cline{2-10}
					& OA (our) &\textbf{37.5851}&\textbf{37.7443} &\textbf{38.4603} &\textbf{38.2592} &\textbf{0.9789}&\textbf{0.9806} &\textbf{0.9816} &\textbf{0.9777}\\		 
					\hline		
					
					\multirow{2}{*}{CO-ImageNet} &SimBA &36.9186 &36.7927 &37.2207 &37.1880 &0.9767 &0.9767 &0.9782 &0.9688  \\
					\cline{2-10}
					& OA (our) &\textbf{37.5211}&\textbf{37.2472} &\textbf{37.9228} &\textbf{38.4209} &\textbf{0.9803}&\textbf{0.9800} &\textbf{0.9823} &\textbf{0.9784}\\					
					\hline
				\end{tabular}
					\begin{tablenotes}
					\footnotesize
					\item Abbreviated by CO-ImageNet: COCO-Reduced-ImageNet.
				\end{tablenotes}
			\end{threeparttable}					
		}
	\end{table*}

\begin{figure*}[!t]	
	\centering
	\includegraphics[width=1\textwidth]{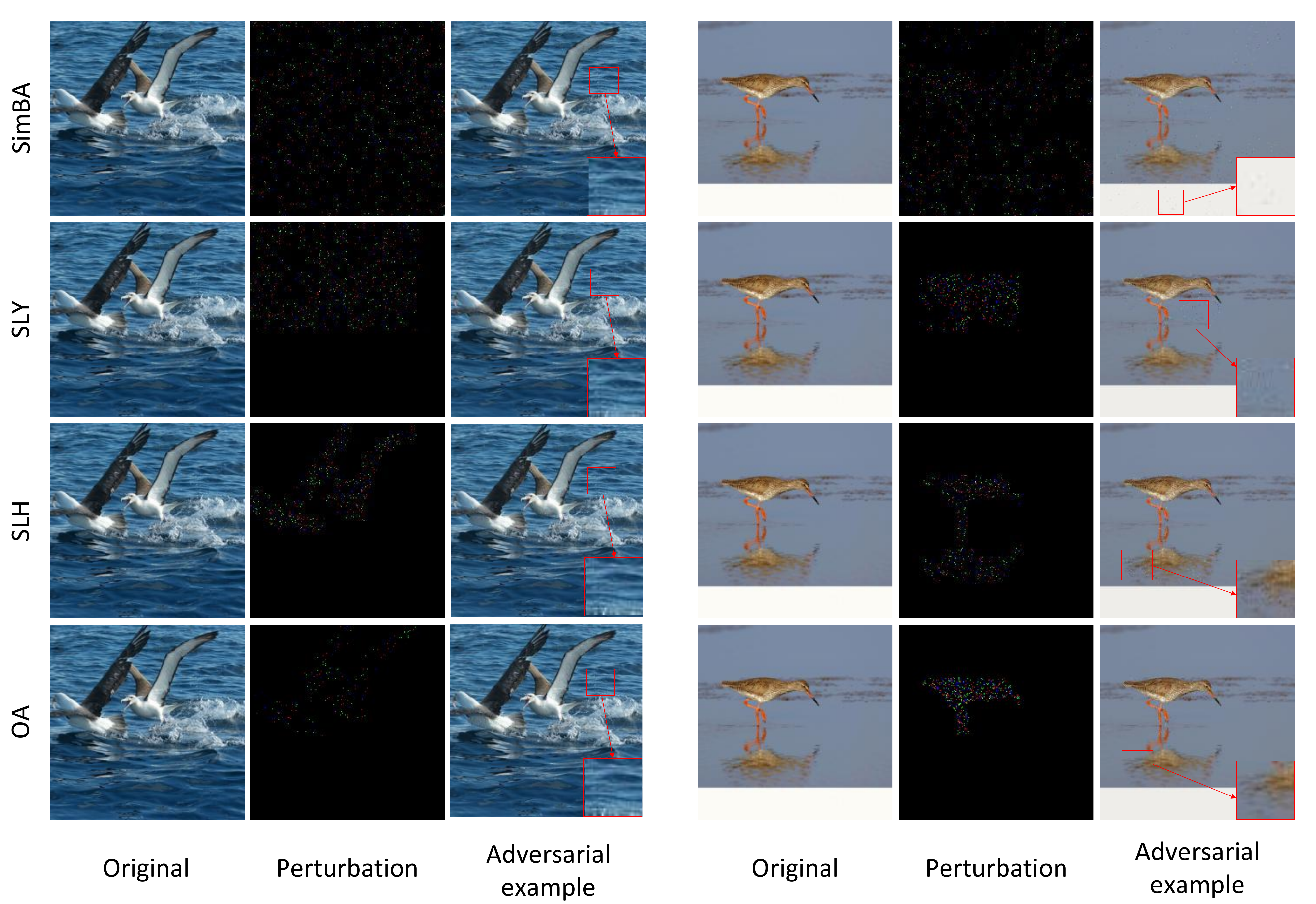}
	\caption{Visual effect of four attack methods on ResNet50. The first and fourth columns
		are the original images, the second and fifth columns are the visual restoration
		of adversarial perturbation, and the third and sixth columns are the adversarial
		examples generated by the corresponding methods. Zoom in to see details.}
	\label{fig1}
\end{figure*}

\section{Conclusion}
In this paper, we have presented an object-attentional untargeted adversarial attack method.  By taking full use of the advantages of YOLOv4 and HVPNet, we proposed a new object segmentation algorithm, avoiding their respective defects. In addition, we designed an activation strategy to avoid the reaction caused by the phenomenon of incomplete detection in SOD. Experimental Results indicate that under variable reasonable settings, our method can not only reduce the number of queries to the target model, but also has better visual hiding effect. 

Note that our model uses the pre-trained weight on COCO 2017 dataset as the network parameter of YOLOv4. Due to the mismatched number of categories between the object detection dataset and image classification dataset, our method has its  limitations: 1) The performance depends on the number of object classes; 2) It is only effective for untargeted attack scenario. Promisingly, researchers would be sure to develop large-scale object detection dataset with much more than 80 categories, which thereby provides an insightful view on our work.




%
%
%
\section{References}
\bibliographystyle{model1-num-names.bst}

\begin{thebibliography}{99}
	\bibitem{1} Y. Wang, et al. AB-FGSM: AdaBelief optimizer and FGSM-based approach to generate adversarial examples, Journal of Information Security and Applications, 68 (2022) 103227.
	\bibitem{2} C. Szegedy, et al. Intriguing properties of neural networks. In: Proceedings of the International Conference on Learning Representations, Banff, Canada, 2014, pp. 1-10.
	\bibitem{3} M. Xue, et al. NaturalAE: natural and robust physical adversarial examples for object detectors, Journal of Information Security and Applications, 57 (2021) 102694.
	\bibitem{4} M. Cheng, et al. Query-efficient hard-label black-box attack: An optimization-based approach. In: Proceedings of the International Conference on Learning Representations, New Orleans, USA, 2019, pp. 1-14.
	\bibitem{5} T. Xiang, et al. Local black-box adversarial attacks: A query efficient approach, arXiv preprint arXiv:2101.01032, 2021.
	\bibitem{6} A. Bochkovskiy, et al. Yolov4: Optimal speed and accuracy of object detection, arXiv preprint arXiv:2004.10934, 2020.
	\bibitem{7} Y. Liu, et al. Lightweight salient object detection via hierarchical visual perception learning, IEEE Transactions on Cybernetics 51 (2020) 4439-4449.
	\bibitem{8} R. Girshick, et al. Region-based convolutional networks for accurate object detection and segmentation, IEEE Transactions on Pattern Analysis and Machine Intelligence 38 (2015) 142-158.
	\bibitem{9} R. Girshick, Fast r-cnn, In: Proceedings of the IEEE International Conference on Computer Vision, Santiago, Chile, 2015, pp. 1440-1448.
	\bibitem{10} S. Ren, et al. Faster r-cnn: Towards real-time object detection with region proposal networks, IEEE Transactions on Pattern Analysis and Machine Intelligence 39 (2016) 1137-1149.
	\bibitem{11} J. Uijlings, et al. Selective search for object recognition, International Journal of Computer Vision 104 (2013) 154-171.
	\bibitem{12} W. Liu, et al. Ssd: Single shot multibox detector. In: Proceedings of the European Conference on Computer Vision, Amsterdam, Netherlands, 2016, pp. 21-37.
	\bibitem{13} S. Chen, et al. Embedding attention and residual network for accurate salient object detection, IEEE Transactions on Cybernetics 50 (2018) 2050-2062.
	\bibitem{14} H. Li, et al. Depthwise nonlocal module for fast salient object detection using a single thread, IEEE Transactions on Cybernetics 51 (2020) 6188-6199.
	\bibitem{15} K. Yan, et al. A new aggregation of DNN sparse and dense labeling for saliency detection, IEEE Transactions on Cybernetics 51 (2020) 5907-5920.
	\bibitem{16} J. Zhao, et al. EGNet: Edge guidance network for salient object detection, In: Proceedings of the IEEE international Conference on Computer Vision, South Korea, 2019, pp. 8779-8788.
	\bibitem{17} M. Kar, et al. A review on progress in semantic image segmentation and its application to medical images, SN Computer Science 2 (2021) 1-30.
	\bibitem{18} Y. Mo, et al. Review the state-of-the-art technologies of semantic segmentation based on deep learning. Neurocomputing 2 (2022) 1-21.
	\bibitem{19} I. Ahmed, M. Jaward, Classifier aided training for semantic segmentation, Journal of Visual Communication and Image Representation, 78 (2021) 103177.
	\bibitem{20} I. Goodfellow, et al. Explaining and harnessing adversarial examples, In: Proceedings of the International Conference on Learning Representations, San Diego, USA, 2015, pp. 1-11.
	\bibitem{21} N. Carlini, D. Wagner, Towards evaluating the robustness of neural networks, In: Proceedings of the IEEE Symposium on Security and Privacy, San Jose, CA, USA, 2017, pp. 39-57.
	\bibitem{22} P. Chen, et al. Zoo: Zeroth order optimization based black-box attacks to deep neural networks without training substitute models, In: Proceedings of the ACM Workshop on Artificial Intelligence and Security, Dallas, Texas, USA, 2017, pp. 15-26.
	\bibitem{23} M. Cheng, et al. Query-efficient hard-label black-box attack: An optimization-based approach, In: Proceedings of the International Conference on Learning Representations, New Orleans, USA, 2019, pp. 1-12.
	\bibitem{24} C. Tu, et al. Autozoom: Autoencoder-based zeroth order optimization method for attacking black-box neural networks, In: Proceedings of the AAAI Conference on Artificial Intelligence, Hawaii, USA, 2019, pp. 742-749.
	\bibitem{25} A. Ilyas, et al. Prior convictions: Black-box adversarial attacks with bandits and priors, In: Proceedings of the International Conference on Learning Representations, New Orleans, Louisiana, USA, 2019, pp. 1-25.
	\bibitem{26} N. Papernot, et al. Practical black-box attacks against machine learning, In: Proceedings of the ACM on Asia Conference on Computer and Communications Security, Abu Dhabi United Arab Emirates, 2017, pp. 506-519.
	\bibitem{27} T. Lin, et al. Microsoft coco: Common objects in context, In: Proceedings of the European Conference on Computer Vision, Zurich, Switzerland, 2014, pp. 740-755.
	\bibitem{28} B. Wandell, J. Winawer. Computational neuroimaging and population receptive fields, Trends in Cognitive Sciences 19 (2015) 349-357.
	\bibitem{29} A. Howard, et al. Mobilenets: Efficient convolutional neural networks for mobile vision applications, arXiv preprint arXiv:1704.04861, 2017
	\bibitem{30} L. Wang, et al. Learning to detect salient objects with image-level supervision, In: Proceedings of the IEEE Conference on Computer Vision and Pattern Recognition, Hawaii, USA, 2017, pp.  136-145.
	\bibitem{31} A. Ilyas, et al. Black-box adversarial attacks with limited queries and information, In: Proceedings of the International Conference on Machine Learning, Chengdu, China, 2018, pp. 2137-2146.
	\bibitem{32} A. Ilyas, et al. Prior convictions: Black-box adversarial attacks with bandits and priors, In: Proceedings of the International Conference on Learning Representations, New Orleans, USA, 2019, pp. 1-25.
	\bibitem{33} C. Guo, et al. Simple black-box adversarial attacks, In: Proceedings of the International Conference on Machine Learning, Long Beach, USA, 2019, pp. 2484-2493.
	
\end{thebibliography}

\end{document}